\documentclass{article}
\usepackage{spconf,amsmath,graphicx}
\usepackage{booktabs}
\usepackage{subcaption}
\usepackage{xcolor}

\title{Physics Informed Generative Models for Magnetic Field Images}
\name{
A.P.P.Aung$^{1\star}$  Lucas Lum$^{2}$  Zhansen Shi$^{2}$  Wen Qiu$^{3}$  Bernice Zee$^{3}$ JM Chin$^{3}$  Yeow Kheng Lim$^{2}$  J.Senthilnath$^{1\star}$
\thanks{$^{\star}$ Corresponding authors: aye\_phyu\_phyu\_aung@i2r.a-star.edu.sg; j\_senthilnath@i2r.a-star.edu.sg.}
\thanks{This study is supported by the Machine Learning Guided Failure Analysis \& Diagnostic Capability Development for Next Generation 3D-IC Packaging at A*STAR via the IAF-PP by the Agency for Science, Technology and Research under Grant No. M23K8a0050.}
}
\address{$^{1}$ Institute for Infocomm Research (I$^{2}$R), A*STAR, Singapore \\
$^{2}$Department of Electrical Engineering, National University of Singapore, Singapore \\
$^{3}$Device Analysis Lab, Advanced Micro Devices (Singapore) Pte Ltd, Singapore\\
}
 
\begin{document}
%\ninept
%
\maketitle
\begin{abstract}
In semiconductor manufacturing, defect detection and localization are critical to ensuring product quality and yield. While X-ray imaging is a reliable non-destructive testing method, it is memory-intensive and time-consuming for large-scale scanning, Magnetic Field Imaging (MFI) offers a more efficient means to localize regions of interest (ROI) for targeted X-ray scanning. However, the limited availability of MFI datasets due to proprietary concerns presents a significant bottleneck for training machine learning (ML) models using MFI. To address this challenge, we consider an ML-driven approach leveraging diffusion models with two physical constraints. We propose \textbf{P}hysics \textbf{I}nformed \textbf{Gen}erative Models for \textbf{M}agnetic \textbf{F}ield \textbf{I}mages (PI-GenMFI) to generate synthetic MFI samples by integrating specific physical information. We generate MFI images for the most common defect types: power shorts. These synthetic images will serve as training data for ML algorithms designed to localize defect areas efficiently. To evaluate  generated MFIs, we compare our model to SOTA generative models from both variational autoencoder (VAE) and diffusion methods. We present a domain expert evaluation to assess the generated samples. In addition, we present qualitative and quantitative evaluation using various metrics used for image generation and signal processing, showing promising results to optimize the defect localization process.

\end{abstract}
\begin{keywords}
Diffusion Models, Physical Constraints, Biot-Savart Law, Generative Models, Magnetic Field Images
\end{keywords}
\section{Introduction}
\label{sec:intro}

In semiconductor manufacturing, defect detection and localization are critical to ensuring product quality and yield. While X-ray imaging is a reliable non-destructive testing method, it is memory-intensive and time-consuming for large-scale scanning due to high resoltion imaging, post processing and precision requirements ~\cite{zeiss2023}. Magnetic Field Imaging (MFI)~\cite{chatraphorn2000scanning} offers a more efficient means to localize regions of interest (ROI) for targeted X-ray scanning. Many physics and electronics research work have proposed several techniques to do MFI based defect localization for defects such as power-to-ground shorts~\cite{vanderlinde2000localizing, mai2001short}, logic failures~\cite{hsiung2004failure} and open circuits~\cite{gaudestad2012space, gaudestad2012spacestacked}. However, the techniques proposed are tailored to specific types of failure analysis and still rely heavily on highly experienced experts, which can significantly slow down the failure analysis process. Therefore, we are motivated to propose machine learning (ML) models to assist engineers in non-destructive failure analysis through MFI-based preliminary defect localization.

However, the limited availability of MFI datasets due to proprietary concerns presents a significant bottleneck for training ML models using MFI for different techniques during the failure analysis process. To address this challenge, we ought to generate MFI datasets for various failure categories to be able to train ML for the techniques tailored to each defect. We consider an ML-driven approach to generate the MFI samples leveraging diffusion models such that they can be used to train the defect localizing ML models for failure analysis pipeline. 

State-of-the-art diffusion models~\cite{rombach2022high, ho2020denoising, bolya2023token} have demonstrated exceptional performance in generating high-quality images across various domains, showcasing their ability to capture intricate details and complex structures. However, these models are primarily trained on trivial image datasets and are not optimized for generating MFIs. When applied to MFIs, diffusion models often fail to produce accurate results because they do not inherently respect the physical constraints dictated by electromagnetic field patterns. These constraints, governed by Maxwell's equations and other principles of electromagnetism, are crucial for maintaining the integrity and realism of magnetic field representations. Without explicitly incorporating these physics-based rules, diffusion models struggle to generate MFIs that align with real-world electromagnetic behavior.

To address these challenges, we make 3 contributions: 

\begin{enumerate}
    \item We propose \textbf{P}hysics \textbf{I}nformed \textbf{Gen}erative Models for \textbf{M}agnetic \textbf{F}ield \textbf{I}mages (PI-GenMFI). We also propose preprocessing, augmentation steps for data scarcity. To the best of our knowledge, we are the first to propose generating an MFI dataset for ML model in non-destructive failure analysis.
    \item We propose formulations of 2 physical constraints inspired by Maxwell Equation and Ampere's Law.
    \item We integrate the two physical constraint formulations along with data preprocessing, augmentation, image enhancing steps to our proposed model as regularization loss to generate MFI images that align with realism of magnetic field representations.
\end{enumerate}

We extensively evaluate our proposed model by using both domain experts' evaluations as well as qualitative and quantitative metrics. We demonstrate the effectiveness of PI-GenMFI in generating MFI that align with electromagnetism rules against SOTA generative model baselines.

\section{Related Works}
\label{sec:relatedworks}
%In this section, we will discuss existing work for failure analysis techniques, generative models, and discussions on their shortcomings.

\subsection{Non-Destructive Failure Analysis}
In non-destructive failure analysis, 3D X-ray~\cite{zulkifli2017high, orozco2018non} is widely used to investigate structural defects with high resolution volumic data.
%like cracks, voids, delamination, and improper solder joints in electronic components. Since 3D-Xray provides high-resolution volumetric data, engineers can analyze both external and internal features without dismantling the device. 
However, high-resolution 3D-Xray scans need 30 minutes to several hours, depending on the sample size and required detail. According to Zeiss' report in  2023~\cite{zeiss2023}, the scanner required 210 minute to image $18 \times 21$ mm TSMC test vehicle sample with suspected failure locations using Xradia 500 Versa microscope. Hence, for more efficient means of failure analysis, research efforts from electric failure analysis propose the use of MFI to analyze current distributions and current crowding in IC chips first, to localize ROI for later targetted X-ray scanning for smaller area~\cite{orozco20133d, gaudestad2014magnetic, orozco2018non}.

Failure analysis by Magnetic Field Imaging (MFI) can be done by many sensors such as Giant Magnetoresistance (GMR)~\cite{baibich1988giant}, Eddy Current Imaging~\cite{garcia2011non}, etc. Among them, Superconducting Quantum Interference Device (SQUID)~\cite{koelle1999high} offers the highest sensitivity, making it ideal for detecting weak currents and deeply buried defects. With SQUID sensor, research has been done for non-destructive 3D-failure analysis for: 1) short localizations single, multi-chip packages~\cite{vanderlinde2000localizing, gaudestad2015magnetic, mai2001short, gaudestad2014short}; 2) high resistance open localizations~\cite{hsiung2004failure, gaudestad2012space, gaudestad2012spacestacked}; 3) failures in daisy chains~\cite{gaudestad2015failure} and even 4) multi-chip failure analysis using 3D reconstructed MFI~\cite{orozco2018non}. However, current research relies on highly skilled physics researchers and engineers to manually analyze MFI results and pinpoint failures, which is a time-consuming process. To address this, we propose utilizing machine learning for more efficient, less labor-intensive failure analysis with MFI. However, we face a bottleneck in obtaining the necessary amount of MFI samples to train a data-driven failure localization ML model due to proprietary restrictions. As a result, before applying ML for defect localization, we explore generating MFI training samples using generative models.

\subsection{Image Generation Models}
Variational Auto-encoder (VAE), Generative Adversarial Networks (GAN) and diffusion models have shown remarkable progress in generating high-quality images of natural images such as faces, scenes, objects, etc. Among these, we did not consider GANs to generate MFI due to training stability and computation efficiency issues~\cite{salimans2016improved, hsieh2019finding, arjovsky2017wasserstein} unlike VAEs and diffusion models. VAEs~\cite{kingma2013auto,burgess2018understanding,van2017neural} apply a probabilistic generating approach with an encoder that maps input data to a latent space and a decoder that generates data from that space. However, VAE suffer from generating blurry images due to the use of Gaussian prior in the latent space. Hence, we mainly consider diffusion models~\cite{sohl2015deep, ho2020denoising, ho2022cascaded, rombach2022high} to generate MFI for their effectiveness to generate non-visual domains~\cite{austin2021structured}. They operate by gradually adding noise to data over a series of steps until it becomes pure noise.
%and then learning to reverse the diffusion process to recover the original data. 
Next, the model learns to predict and remove the added noise at each step through a neural network, and the final generative process involves sampling from pure noise and progressively denoising it back to a high-quality image. SOTA models such as Cascaded Diffusion Models~\cite{ho2022cascaded} and Latent Diffusion Models~\cite{rombach2022high} compressed latent space for faster inference times. However, we need a more complex encoder-decoder  design to correctly represent MFI samples in the latent space. Furthermore, adding physical constraints in the latent space is more challenging compared to working with pixel representations, as the latent space may not directly correspond to the physical properties that need to be modeled. 

On the other hand, a slightly older diffusion model, DDPM~\cite{ho2020denoising} operates in the feature space. Hence, we can easily formulate and apply physical constraints to DDPM's objective function as they are calculated and enforced at the pixel level in the electromagnetic physics context. Since DDPMs generate images by iteratively denoising them, physical properties or domain-specific constraints can be more directly incorporated into the pixel space during the training and generation process.

\section{PI-GenMFI}

\begin{figure}[ht]
    \centering
    \includegraphics[width=\linewidth]{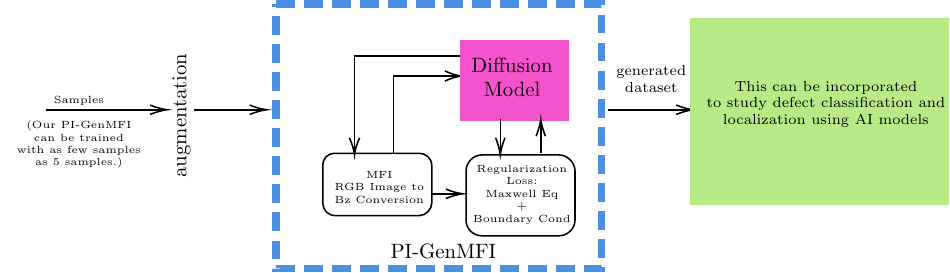}
    \caption{Overview of PI-GenMFI Architecture}
    \label{fig:architecture}
\end{figure}

Fig.~\ref{fig:architecture} describes the overview of the PI-GenMFI architecture, which is built on established generative models such as Denoising Diffusion Probabilistic Models (DDPM) [2]. In simple terms, DDPM starts by corrupting an image or dataset with increasing levels of noise, eventually turning it into pure noise. During training, the model learns how to denoise the data at each step of the process. By reversing this, the model can generate new samples from random noise, producing realistic images or data. We have gathered both real and simulated datasets (a total of 3 samples, with (4,4,9) configurations each.) from our industry collaborators to train our proposed PI-GenMFI. To adapt DDPM for training with MFI samples, we applied a series of augmentation and preprocessing steps, as well as image enhancement to ensure that the generated images closely resemble the characteristics of MFI data.

\subsection{Data Preparation and Augmentation}

\begin{figure}[h!]
    \centering
    \begin{minipage}{0.15\textwidth}
        \centering
        \includegraphics[width=\linewidth]{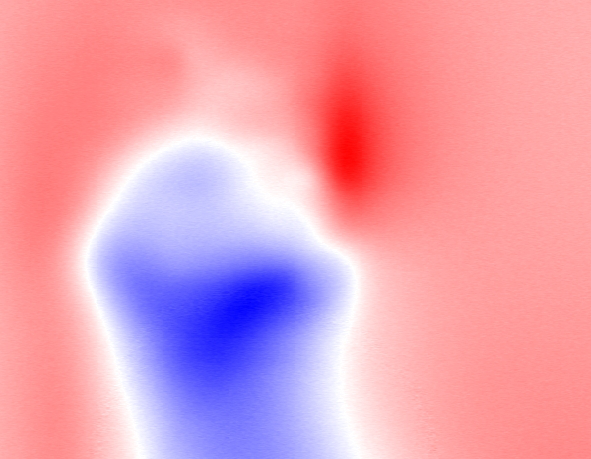}
        \subcaption{Expert 1}
    \end{minipage}
    \begin{minipage}{0.3\textwidth}
        \centering
        \includegraphics[width=\linewidth]{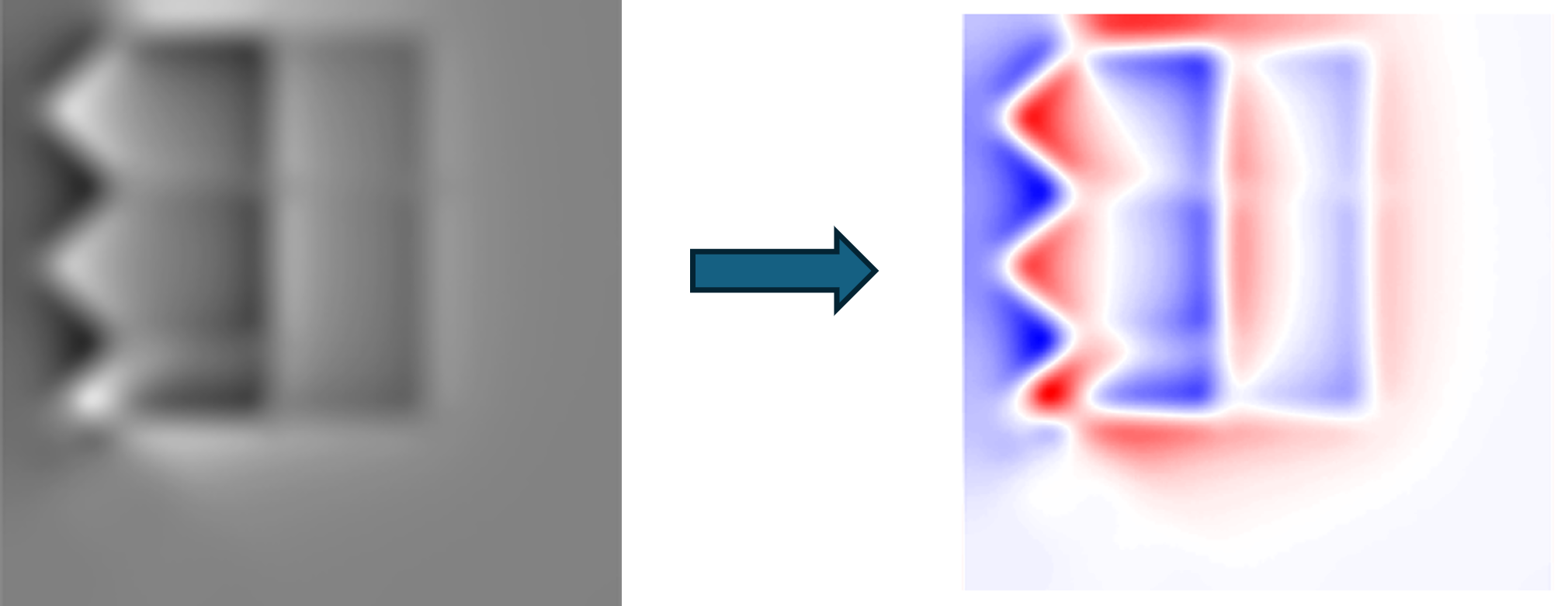}
        \subcaption{Expert 2}
    \end{minipage}
    \caption{Visualizing and preprocessing of data samples from collaborators}
    \label{fig:preprocess}
\end{figure}

\noindent \textbf{Preprocessing.} We obtained a limited set of MFI samples from two of our collaborators. However, these samples are not directly suitable for training image generation models. To address this, we normalized the original image colors (ranging from 0 to 255) using the Bz values provided in the metadata and replotted the MFI using the cmap.bwr color map. The changes made during our data preparation process are shown in the Fig.~\ref{fig:preprocess}.

\noindent \textbf{Augmentation.}
Given the limitation of only 17 MFI samples, training a diffusion model directly is infeasible. To address this, we generated a training dataset by introducing Gaussian noise, pink noise, and applying transformations such as rotation and warp shifting. These preprocessing steps enhance the diversity and robustness of the dataset, facilitating an improved model training process.

\subsection{Adding Physical Constraints} 
To ensure that the generated Magnetic Field Images (MFI) from PI-GenMFI closely resemble the scanned or simulated MFI, we propose the physical constraints to integrate into the diffusion process. Specifically, we introduce two electromagnetism-related principles to guide the model. %Maxwell's Equations and Boundary Condition.

\noindent \textbf{Physical Constraint 1: Gauss's Law.} Gauss' Law for Magnetism~\cite{maxwell1865viii} ensures the charge conservancy of the magnetic fields. It states that the divergence of magnetic flux across any closed surface should be zero,  s.t., $
\nabla \cdot \mathbf{B} = 0$. Since the isolated magnetic poles (monopoles) do not exist, the first physical constraint,  $\mathcal{L}_1 = |\nabla \cdot \mathbf{B}|$, ensures that the generated MFIs do not have the magnetic field lines that do not form closed loops. The magnetic field $\mathbf{B}$ in 2D is:

\vspace{-0.2cm}
\small
\begin{equation}
    \mathbf{B} (x,y) = B_x(x, y) \, \hat{i} + B_y(x, y) \, \hat{j},
\end{equation}
\normalsize

where $B_x(x, y)$ and $B_y (x,y)$ are magnetic field component in $x, y$ direction. $\hat{i}$ and $\hat{j}$ are directional vectors. Next, we calculate the divergence of the magnetic field of a closed surface by partial derivatives with respect to the coordinates:

\vspace{-0.2cm}
\small
\begin{equation}
 \nabla \cdot \mathbf{B} = \frac{\partial B_x}{\partial x} + \frac{\partial B_y}{\partial y}   
\end{equation}
\normalsize

% \begin{multline}
%     \nabla \cdot \mathbf{B} \approx \frac{f_x(i+1, j) - f_x(i-1, j)}{2 \Delta x} \\ 
%   + \frac{f_y(i, j+1) - f_y(i, j-1)}{2 \Delta y}
% \end{multline}

% Hence,We add it as a constraint to our PI-GenMFI training ensures that the inconsistent magnetic field strengths that do not correlate with the current distributions are not generated. 
\noindent \textbf{Physical Constraint 2: Boundary Condition.} We derive the second from Ampere's Law~\cite{griffiths2023introduction} which states that the strength of the magnetic field is directly proportional to the magnitude of the current. Hence, we add the standard boundary condition, i.e., the magnetic field approaches zero at infinity to reflect the fact that the magnetic field generated by an electromagnetic system must eventually diminish as we move away from the source. Integrating the boundary condition to our PI-GenMFI training captures the phenomenon of decaying of magnetic field to avoid generation of unrealistic magnetic field patterns at the boundary. The boundary condition is defined as: 

\vspace{-0.2cm}
\small
\begin{equation}
    \mathbf{B}(x,y) = 0, \text{ for } x=0, x=L_{x}, y =0, y=L_{y}, 
\end{equation}
\normalsize

where $x \in [0, L_{x}]$ and $y \in [0,L_{y}]$. Hence, we define the second physical constraint as $\mathcal{L}_2 = |\mathbf{B}(x,y)|$.

\subsection{Regularization Step}
We add the two additional loss terms defined by the physical constraints to the DDPM's MSE loss as regularization terms. The training objective of PI-GenMFI is given by:

\vspace{-0.2cm}
\small
\begin{equation}
    \mathtt{L} = \mathcal{L}_{MSE} + \alpha \cdot \mathcal{L}_{1} + \beta \cdot \mathcal{L}_{2},
\end{equation}

\noindent where $\alpha=0.5$ and $\beta=0.2$ are hyperparameters to set the contribution of physical losses to the objective function. 

\normalsize
 
\subsection{Image Enhancing of PI-GenMFI Results}
To make the PI-GenMFI generated images better resemble the original samples in terms of saturation, brightness and contrast, we increased saturation = 2.0, brightness = 1.5 and contrast = 1.5 using the Pillow~\cite{Pillow} package. 

\section{Evaluation}
\subsection{Baselines}
We compare our proposed PI-GenMFI with baselines such as Variational Autoencoder (VAE), VAE with physics constraints (VAE+Phys) and Denoising Diffusion Probabilistic Model (DDPM). Lastly, we add 2 versions of our proposed model: PI-GenMFI and PI-GenMFI with image enhancement (PI-GenMFI-eh).

\begin{figure}[ht]
    \centering
    % First row of subfigures
    \begin{subfigure}[b]{0.23\textwidth}
        \centering
        \includegraphics[width=\textwidth]{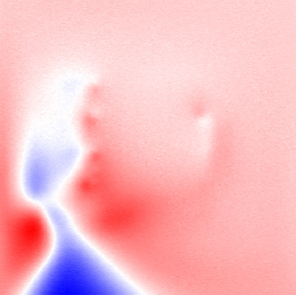}
        \caption{Original}
        \label{fig:sub1}
    \end{subfigure}
    \hfill
    \begin{subfigure}[b]{0.23\textwidth}
        \centering
        \includegraphics[width=\textwidth]{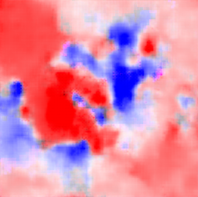}
        \caption{VAE}
        \label{fig:sub2}
    \end{subfigure}
     \vspace{1em} % Space between rows
     
    \begin{subfigure}[b]{0.23\textwidth}
        \centering
        \includegraphics[width=\textwidth]{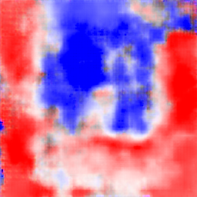}
        \caption{VAE+Phys}
        \label{fig:sub3}
    \end{subfigure}
   \hfill
    % Second row of subfigures
    \begin{subfigure}[b]{0.23\textwidth}
        \centering
        \includegraphics[width=\textwidth]{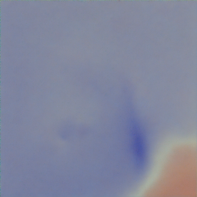}
        \caption{DDPM}
        \label{fig:sub4}
    \end{subfigure}
     \vspace{1em} % Space between rows
    \begin{subfigure}[b]{0.23\textwidth}
        \centering
        \includegraphics[width=\textwidth]{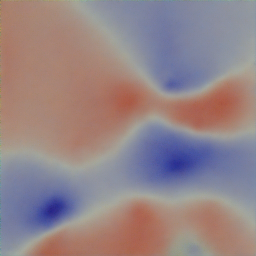}
        \caption{PI-GenMFI}
        \label{fig:sub5}
    \end{subfigure}
    \hfill
    \begin{subfigure}[b]{0.23\textwidth}
        \centering
        \includegraphics[width=\textwidth]{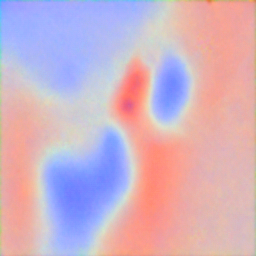}
        \caption{PI-GenMFI-eh}
        \label{fig:sub6}
    \end{subfigure}

    \caption{Comparison of Generated MFI from our PI-GenMFI against original samples and results from baselines.}
    \label{fig:genresults}
\end{figure}

\subsection{Training Data}
We received two real IC samples with intra-plane power short defects from Expert 1. For each sample, we collected four MFI scans, each corresponding to a different pin configuration. Additionally, we received one simulated IC sample with an inter-plane short defect from Expert 2. For this sample, we obtained nine MFI scans, including three pin configurations and three offset configurations.

\subsection{Evaluation Metrics}
For all metrics, $I(x,y)$ represents the original image and $G(x,y)$ the generated image.

\noindent \textbf{Average PSNR.} Peak Signal-to-Noise Ratio compares the similarity between an original and the generated image. 

\vspace{-0.2cm}
\small
\begin{align}
\text{PSNR} &= 10 \cdot \log_{10} \left( \frac{I_{\text{max}}^2}{\text{MSE}} \right),  \\
\text{MSE} &= \frac{1}{m \times n} \sum_{x=1}^{m} \sum_{y=1}^{n} \left( I(x, y) - G(x, y) \right)^2.
\end{align}
\normalsize

\noindent \textbf{Average SSIM.} Structural Similarity Index measures the similarity between two images, considering luminance, contrast, and structural differences. SSIM ranges from $[-1,1]$ where $1$ indicates perfect similarity between the two images. 

\vspace{-0.2cm}
\small
\begin{equation}
\text{SSIM}(I, G) = \frac{(2 \mu_I \mu_G + C_1)(2 \sigma_{IG} + C_2)}{(\mu_I^2 + \mu_G^2 + C_1)(\sigma_I^2 + \sigma_G^2 + C_2)},
\end{equation}
\normalsize

where $\mu$ is the mean intensity. $\sigma$ is the variance. $\sigma_{IG}$ is the covariance between images $I$ and $G$. $ C_1$ and $ C_2$ are small constants to stabilize the division.

\noindent \textbf{FID.} Fréchet Inception Distance evaluates the quality of images generated by generative models. We let $p$ and $q$ be the distributions of the representations obtained by projecting $I$ and $G$ to the last hidden layer of a pretrained Inception model. Assuming multivariate Gaussian distributions $p \sim \mathcal{N}(\mu_p, \Sigma_p)$ and $q \sim \mathcal{N}(\mu_q, \Sigma_q)$ where $\mu$ is the mean and $\Sigma$ is the variance, FID measures the 2-Wasserstein distance between $p$ and $q$.

\vspace{-0.3cm}
\small
\begin{multline}
    \text{FID}(p, q) = W_2(p, q) = \| \mu_p - \mu_q \|^2 +  \\ \text{Tr} \left( \Sigma_p + \Sigma_q - 2 \left( \Sigma_p \Sigma_q \right)^{\frac{1}{2}} \right).
\end{multline}
\normalsize

\noindent \textbf{Fourier Score.} Fourier Score (FS) measures how well the generated images capture the frequency information of real images. We first compute the 2D Fourier Transform $F(.)$ of the original and generated image, 
then calculate the magnitude spectrum and lastly, measure the similarity. 

\vspace{-0.2cm}
\small
\begin{equation}
\text{Fourier Score}(I, G) = \frac{\| F(I) - F(G) \|_2^2}{\| F(I) \|_2^2 + \| F(G) \|_2^2}.
\end{equation}
\normalsize

\noindent \textbf{Divergence Score.} We reuse Physical Constraint 1 in Section 3.2. for this metric. %Specifically, for an MFI sample $B(x,y)$, we define the divergence score $\nabla \cdot \mathbf{B}$ as:
% %\vspace{-0.2cm}
% \small
% \begin{multline}
%     \nabla \cdot \mathbf{B} \approx \frac{f_x(i+1, j) - f_x(i-1, j)}{2 \Delta x} \\ 
%   + \frac{f_y(i, j+1) - f_y(i, j-1)}{2 \Delta y}
% \end{multline}
% \normalsize

\begin{table}[ht]
\centering
\caption{Domain expert evaluation.}
\normalsize
\begin{tabular}{lccc}
\toprule
                      & Samples & Expert 1 & Expert 2 \\
                      \midrule
VAE                   & 20      & 25\%            & 0\%             \\
VAE+Phys              & 20      & 15\%            & 0\%             \\
DDPM                  & 20      & 55\%            & 0\%             \\
PI-GenMFI & 20      & \underline{70}\%            & \textbf{65\%}            \\
PI-GenMFI-eh             & 40      & \textbf{80\%}           & 0\%  \\          \bottomrule
\end{tabular}
\label{experttable}
\end{table}
\begin{table*}[ht]
\centering
\caption{Comparison of generated powershort MFIs using metrics from both signal processing and generative ML.}
\normalsize
\begin{tabular}{lc|cc|ccc}
\toprule
     & Original & VAE      & VAE+Phys & DDPM    & PI-GenMFI & PI-GenMFI-eh \\ 
     \midrule
PSNR ($\uparrow$) &     -     & 11.564   & 11.313   & 10.772  & 10.045    & \textbf{16.847}             \\
SSIM ($\uparrow$) &     -     & 0.48     & 0.462    & 0.686   & \underline{0.712}     & \textbf{0.717}              \\
FID  ($\downarrow$) &     -     & 1529.024 & 1537.39  & 1240.28 & \textbf{1122.84} & 1440.705      \\
FS  ($\uparrow$) & 1923.112 & 1734.25  & 1711.33  & 592.78  & 397.25    & \textbf{763.87}             \\
Div ($\downarrow$) & 0.0045   & 0.129   & 0.106    & 0.124   & \underline{0.103}     & \textbf{0.0205} \\
\bottomrule
\end{tabular}
\label{resultsTable}
\end{table*}

\subsection{Domain Expert Evaluation}

To verify the generated datasets from all the models, we sent the anonymized generated samples from all 5 generative models to two domain expert teams to assess their resemblance to real samples. We invited Expert 1 from industry and Expert 2 from academic collaborators. Table~\ref{experttable} presents the generated sample counts that we provided to each expert along with their respective evaluations.

The results indicate that both versions of our proposed generative models outperform the baselines in terms of expert acceptance. Despite the outperformance in terms of evaluation metrics, the detailed feedback from Expert 2 highlighted that the generated samples from PI-GenMFI-eh appear oversaturated, leading them to rate it as unrealistic. In contrast, our industry partner Expert 1 found it acceptable and useful. Consequently, we conclude that PI-GenMFI is capable of adhering to the physics rules and maintaining the electrical patterns in the generated samples.

\subsection{Qualitative Evaluation}

Fig.~\ref{fig:genresults} shows the qualitative comparison of the generated images our PI-GenMFI against the baselines. The results from both the VAE and VAE+Phys models were blurry, confirming the general understanding that VAE-based models struggle with generating sharp images. Additionally, our experiment results revealed that directly applying DDPM performed poorly in generating MFI samples, both in terms of adhering to electromagnetism rules and maintaining appropriate color contrast. To address these limitations, we incorporated preprocessing and augmentation techniques before training and applied image enhancement after generation. These adjustments allowed our proposed PI-GenMFI models to generate MFI samples that closely resemble the original samples, adhere to electromagnetism rules, and exhibit diverse and realistic patterns.

\subsection{Quantitative Evaluation}

Table~\ref{resultsTable} presents a quantitative comparison of generated MFIs using various metrics from signal processing and generative machine learning (ML) approaches. First, we observe that the SSIM values for both VAE (0.48) and VAE+Phys (0.462) are relatively low, indicating that these models struggle to maintain the structural integrity of the generated MFIs compared to the original. We can conclude that VAEs, even with physics constraints, are not as effective as diffusion models (DDPM=0.686) in capturing the complex structural patterns inherent in MFIs. 

Next, we observe that PI-GenMFI-eh outperforms all other diffusion models in most metrics, including PSNR, SSIM, Fourier Score (FS), and Divergence Score (Div). It achieves the highest PSNR (16.847), SSIM (0.717), and FS (763.87), indicating superior fidelity, structural consistency, and alignment with the original MFIs. It also has the lowest Divergence Score (0.0205), reflecting excellent control over the divergence from expected electromagnetic patterns.

PI-GenMFI mostly tails as a close second while performing the best (1122.84) on FID. However, we find that FID is less reliable in this context, as it is typically used for natural images, where scores below 10 are considered benchmarks for high-quality results. Since MFIs are specialized images that differ significantly from natural image datasets, FID may not provide an accurate measure of perceptual quality.

\iffalse
\section{Future Directions}
Since this is the initial step in generating MFI simulations using generative AI, the preliminary results show some limitations compared to real-life defect detection scenarios. These limitations include the generation of multiple defect areas and the presence of multiple root causes within a single MFI image. To address these challenges, we plan to further explore the ability to guide the generation process using text prompts. Specifically, we aim to define key parameters such as the number of defects in an MFI image, the root cause (whether inter-plane or infra-plane), and input markings to indicate the location of current for more accurate defect localization. This approach is intended to enhance the relevance and precision of the generated MFI images in real-world applications.
\fi

\section{Conclusion}

We introduced a novel idea to enable non-destructive ML-guided IC chip defect analysis, called Physics Informed Generative Models for Magnetic Field Images (PI-GenMFI), incorporating data preprocessing, augmentation, and two physical constraint formulations based on Maxwell's Equation and Ampere's Law, integrated as regularization loss to enhance the realism of MFI generation. Extensive evaluations demonstrate that PI-GenMFI outperforms state-of-the-art models in producing MFI images that align with electromagnetism principles. Since this is initial step in generating MFI using generative AI, preliminary results still have some limitations, such as the generation of multiple defects and root causes within a single image. In the future works, we plan to explore guiding the generation process with text prompts to define parameters such as defect count, root cause, and current source, aiming to improve the relevance of generated MFI images for real-world applications. Moreover, our proposed method can also be extended to other modes of failure analysis scannings such as X-rays and thermal imaging.

\newpage
\small
\bibliographystyle{IEEEbib}
\bibliography{refs}

\end{document}